\documentclass{article}

\usepackage{PRIMEarxiv}

\usepackage[utf8]{inputenc} % allow utf-8 input
\usepackage[T1]{fontenc}    % use 8-bit T1 fonts
\usepackage{hyperref}       % hyperlinks
\usepackage{url}            % simple URL typesetting
\usepackage{booktabs}       % professional-quality tables
\usepackage{amsfonts}       % blackboard math symbols
\usepackage{amsmath}        % added for advanced math environments
\usepackage{nicefrac}       % compact symbols for 1/2, etc.
\usepackage{microtype}      % microtypography
\usepackage{lipsum}
\usepackage{fancyhdr}       % header
\usepackage{graphicx}       % graphics
\graphicspath{{media/}}     % organize your images and other figures under media/ folder

\title{From Natural Language to Real Schematics: An Automated Workflow for Power Electronics Schematic Synthesis
}

\author{
  Bowen Su, Huan Chen, Jiahua Ying, Ran Ou, Weilong Chen, Li Xue, Kai Sun\thanks{\textbf{Corresponding Author}} \\
  State Key Laboratory of Power System Operation and Control\\
  Department of Electrical Engineering \\
  Tsinghua University \\
  Beijing, China\\
  \texttt{subw24@mails.tsinghua.edu.cn, sun-kai@mail.tsinghua.edu.cn}
}

\begin{document}
\maketitle

\begin{abstract}
Designing power electronics converters traditionally requires intensive human expertise across topology sizing, commercial component selection, circuit simulation verification, and schematic drafting. While Large Language Models (LLMs) show promise in electronic design automation (EDA), applying standalone LLMs directly to board-level power electronics schematic synthesis suffers from severe engineering hallucinations—including inaccurate analytical sizing, fabricated or mismatched component part numbers, violated voltage derating margins, and broken topological connectivity. In this paper, we propose a Bounded Dual-Loop Automated Schematic Synthesis Framework that translates natural language specifications into verified, editable KiCad schematics and manufacturing deliverables. Our workflow integrates six stages across three functional columns: (1) requirement parsing and knowledge-driven power stage synthesis covering non-isolated (Buck, Boost) and isolated/resonant (Flyback, DAB, half-bridge LLC) topologies; (2) contract- and evidence-coordinated component selection grounded in distributor APIs and OEM datasheets, coupled with closed-loop \texttt{ngspice} transient verification and stress feedback; and (3) modular control hardware integration and multi-format deliverable generation featuring native KiCad netlist round-trip verification. End-to-end case studies on a $24\text{--}36\,\text{V}$ to $12\,\text{V}/2\,\text{A}$ Buck converter (29 positions, 83 verified terminals) and an isolated $48\,\text{V}$ to $24\,\text{V}/0.5\,\text{A}$ Dual-Active Bridge (DAB) converter (87 positions, 341 verified terminals), alongside comparative evaluations against a Vanilla LLM baseline, demonstrate that our framework improves component verification coverage from $22.2\%$ to $98.6\%$, nominal value consistency from $0.0\%$ to $97.7\%$, and SPICE simulation pass rates from $0.0\%$ to $100.0\%$.
\end{abstract}

% keywords

% ==========================================
% Section 1: Introduction
% ==========================================
\section{Introduction}
\label{sec:intro}

% Paragraph 1: 电力电子的大背景与快速定制化需求
Power electronics serves as the foundational backbone of modern electrical energy conversion and management, playing a pivotal role in renewable energy integration, electric vehicles (EVs), industrial motor drives, and next-generation artificial intelligence (AI) data center power delivery networks, as shown in Figure 1. With the rapid proliferation of wide-bandgap (WBG) semiconductors—such as Silicon Carbide (SiC) and Gallium Nitride (GaN)—and increasingly diverse application scenarios, the industry faces an urgent demand for customized power converters spanning a wide range of voltage and power ratings. Accelerating the hardware development lifecycle and achieving rapid prototyping from high-level system specifications have become critical competitive advantages in modern power electronics engineering.

% Paragraph 2: 电力电子原理图设计的重要性与传统人工流程的繁琐
Within the power electronics development lifecycle, circuit schematic design stands as the most critical link bridging abstract functional requirements and physical hardware implementation. A complete engineering schematic not only defines the circuit topology and connectivity but also dictates specific commercial component selections, electrical and thermal stress margins, and downstream printed circuit board (PCB) layout constraints. Traditionally, translating a functional requirement (e.g., converting a 500V DC bus to a regulated 200V output) into a production-ready schematic is a labor-intensive, iterative process heavily reliant on experienced human engineers. Engineers must manually select appropriate topologies, derive steady-state and dynamic equations to size passive components, comb through massive distributor catalogs and datasheets to select discrete semiconductors, construct simulation models to verify waveforms, and finally compile engineering deliverables such as the Bill of Materials (BOM) and netlists. This high barrier to entry and prolonged iteration cycle severely bottleneck design productivity.
\begin{figure}
    \centering
    \includegraphics[width=0.5\linewidth]{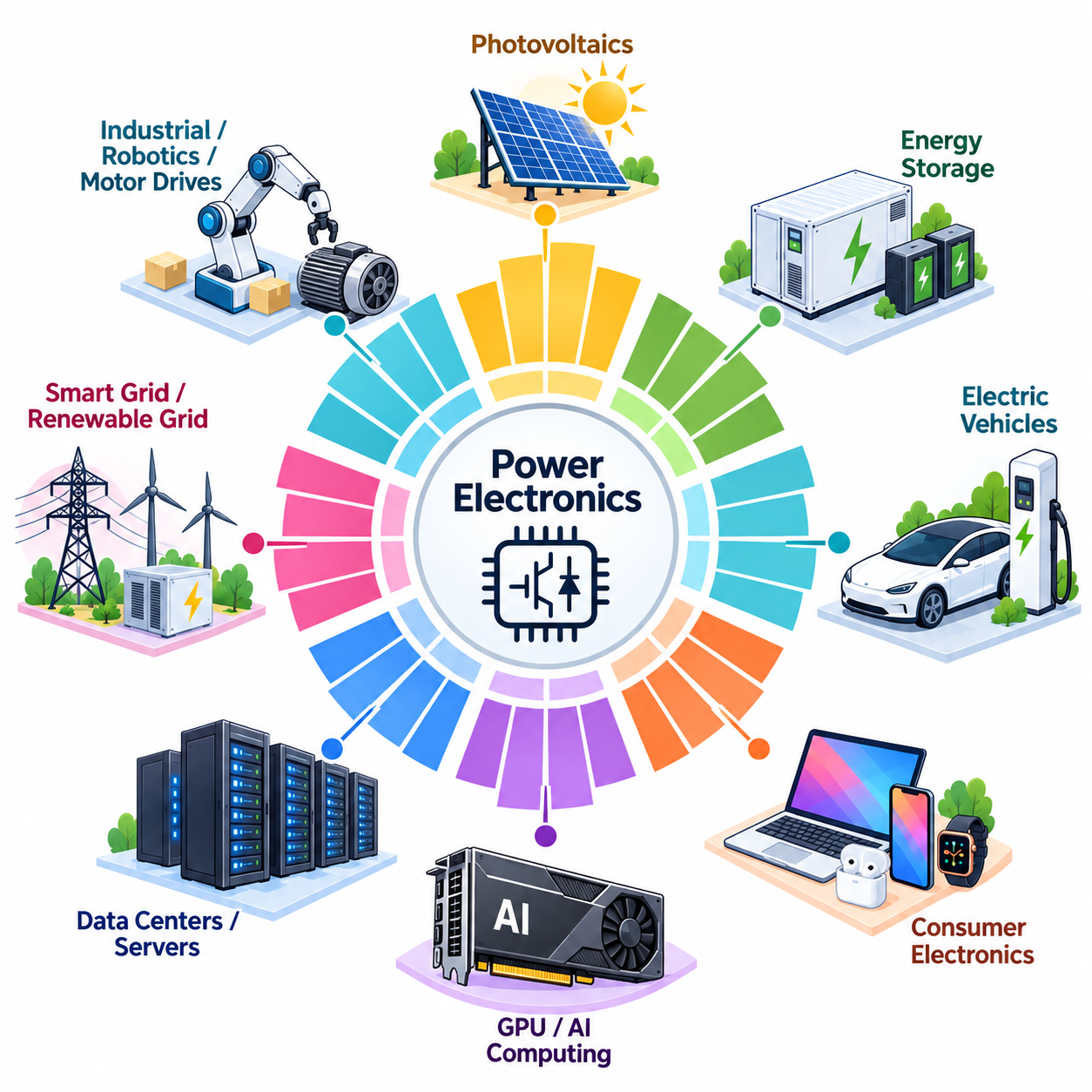}
    \caption{Application scenarios of power electronics}
    \label{fig:placeholder}
\end{figure}
% Paragraph 3: 传统自动化工具的局限 & 单个 LLM 面临的三大痛点（幻觉、算不准、无工程经验/无闭环验证）
While early electronic design automation (EDA) efforts in power electronics introduced expert systems and metaheuristic optimization algorithms (e.g., genetic algorithms and particle swarm optimization) to assist in parameter sizing, these rigid frameworks require predefined mathematical formulations for fixed topologies and lack the flexibility to process natural language intent or generate comprehensive engineering documentation. Recently, Large Language Models (LLMs) have demonstrated remarkable capabilities in software engineering and digital integrated circuit design (such as automated Verilog generation). However, directly applying a standalone, monolithic LLM to power electronics schematic design encounters severe practical bottlenecks:
\begin{enumerate}
    \item \textbf{Inaccurate Analytical Computation:} As probabilistic token predictors, standalone LLMs frequently fail at precise numerical calculations required for power stage sizing (e.g., inductance, capacitance, and root-mean-square current derivations), leading to physically mismatched circuit parameters.
    \item \textbf{Component Hallucination and Lack of Engineering Experience:} Vanilla LLMs lack real-time access to commercial supply chains and practical engineering heuristics. Consequently, they often fabricate non-existent part numbers or neglect critical safety derating rules (e.g., selecting a 500V-rated MOSFET for a 500V bus without accounting for switching voltage spikes), which would result in immediate catastrophic failure in real hardware.
    \item \textbf{Open-Loop Generation without Physical Verification:} Generating a circuit schematic or netlist in a single pass without circuit simulator feedback provides no guarantee of electrical functionality, often yielding syntax-broken netlists, non-convergent simulations, or severe deviations in steady-state output voltage and ripple specifications.
\end{enumerate}

% Paragraph 4: 我们的工作（Workflow 核心设计）
To overcome these challenges, we propose an automated, end-to-end agentic workflow that synthesizes verified power electronics schematics and complete engineering deliverables directly from natural language prompts (e.g., \textit{``Design a 500V-to-200V Buck converter''}). Rather than relying on a single LLM's internal parametric memory, our framework orchestrates the LLM with specialized external tools and domain-specific engineering rules. Specifically, the workflow parses user requirements and invokes programmatic calculation modules embedded with practical engineering derating margins to derive exact theoretical component stresses. Subsequently, it bridges the gap between theoretical specs and real-world hardware by querying commercial electronic component distributor APIs to select authentic, in-stock devices. Furthermore, to ensure electrical reliability, the workflow constructs an executable SPICE netlist and establishes a closed-loop verification mechanism with NGSPICE. By automatically analyzing simulated time-domain waveforms against design targets, the agent iteratively refines circuit parameters until full functional compliance is achieved, ultimately outputting the finalized schematic, BOM, netlist, and design documentation.

% Paragraph 5: 核心贡献总结
The primary contributions of this paper are summarized as follows:
\begin{itemize}
    \item \textbf{An End-to-End Schematic Synthesis Workflow:} We develop a novel LLM-based agentic architecture tailored for power electronics engineering, automating the entire pipeline from natural language requirement parsing to the generation of complete schematics, BOMs, SPICE netlists, and design specifications.
    \item \textbf{API-Grounded Component Selection with Engineering Heuristics:} We eliminate LLM component hallucinations by integrating real-world distributor APIs with domain-specific voltage/current derating rules, ensuring that all synthesized schematics utilize commercially available and safely rated power devices.
    \item \textbf{Closed-Loop NGSPICE Verification:} We introduce a simulation-in-the-loop self-correction mechanism that validates generated circuits via NGSPICE transient simulations, enabling automated diagnosis and iterative tuning of circuit parameters to guarantee steady-state and ripple performance.
\end{itemize}

% Paragraph 6: 全文行文安排 (Paper Organization)
The remainder of this paper is organized as follows. Section~\ref{sec:related} reviews related work in automated power electronics design and LLM-aided hardware synthesis. Section~\ref{sec:method} details the system architecture and the four core stages of the proposed automated workflow. Section~\ref{sec:experiments} presents an end-to-end engineering case study on a 500V-to-200V converter alongside ablation evaluations. Section~\ref{sec:discussion} discusses current limitations and future extensions, and Section~\ref{sec:conclusion} concludes the paper.

% ==========================================
% Section 2: Related Work
% ==========================================
\section{Related Work}
\label{sec:related}

\subsection{Design Automation and AI in Power Electronics}
Automating the design of power electronics converters has been a long-standing pursuit to reduce engineering development cycles and optimize system performance~\cite{hermanns2020increasing}. As comprehensively reviewed by Zhao et al.~\cite{zhao2021overview}, traditional artificial intelligence (AI) and computer-aided design (CAD) techniques in power electronics have primarily relied on expert systems, metaheuristic algorithms (e.g. genetic algorithms and particle swarm optimization), and data-driven surrogate models to optimize converter efficiency, weight, or thermal performance~\cite{li2022artificial}. In terms of physical and module-level automation, tools such as PowerSynth~\cite{alrazi2021powersynth} have pioneered layout synthesis and electro-thermal optimization for multichip power modules. 

Recently, inspired by the reasoning capabilities of foundation models, researchers have begun exploring Large Language Models (LLMs) for power electronics applications. Notably, Lin et al. proposed PE-GPT~\cite{lin2025pegpt}, a multimodal LLM framework enhanced with retrieval-augmented generation (RAG) and metaheuristic algorithms to assist in dual-active bridge (DAB) modulation design and converter parameter sizing. Similarly, Guo et al.~\cite{guo2026accessible} demonstrated a customized GPT framework for photovoltaic boost converter design and verification. However, these pioneering efforts primarily focus on theoretical parameter optimization or modulation strategy selection. They do not address the automated synthesis of complete engineering schematics grounded in real-time commercial component supply chains, nor do they automate the generation of manufacturing deliverables such as BOMs and verified netlists.

\subsection{Large Language Models for Hardware and Circuit EDA}
In the broader Electronic Design Automation (EDA) community, LLMs have catalyzed a paradigm shift in hardware synthesis~\cite{zhong2023llm4eda}. Initially, research efforts concentrated on digital integrated circuit (IC) design, leveraging domain-adapted LLMs to generate Register-Transfer Level (RTL) Verilog code and EDA tool scripts, exemplified by NVIDIA's ChipNeMo~\cite{liu2023chipnemo} and VerilogEval~\cite{liu2023verilogeval}. 

More recently, attention has expanded from digital logic to analog circuit topology generation and SPICE netlist synthesis. For instance, AnalogCoder~\cite{lai2025analogcoder} introduced a training-free LLM agent that generates Python-based PySpice code to design analog amplifiers via feedback-driven prompts. Similarly, SPICEPilot~\cite{vungarala2024spicepilot} and AnalogXpert~\cite{zhang2024analogxpert} explored LLM-guided SPICE code generation and subcircuit-level topology synthesis. Despite these advances, existing circuit-level LLM frameworks target low-voltage, integrated CMOS analog circuits, where design degrees of freedom are largely confined to transistor channel width-to-length ($W/L$) sizing. In contrast, board-level power electronics schematic design operates under high-voltage and high-current switching regimes. It requires selecting discrete commercial off-the-shelf (COTS) power semiconductors (e.g., SiC MOSFETs, fast-recovery diodes) and passive components with strict voltage/current derating margins—a critical engineering dimension absent in IC-oriented LLM frameworks.

\subsection{Tool-Augmented and Verification-in-the-Loop Agents}
To mitigate the inherent limitations of standalone LLMs in mathematical calculation and factual accuracy, the AI community has developed agentic paradigms that interleave reasoning with external tool execution, such as ReAct~\cite{yao2023react} and Toolformer~\cite{schick2023toolformer}. By delegating arithmetic operations to symbolic interpreters and querying external APIs, tool-augmented agents significantly curb hallucinations. Furthermore, incorporating domain-specific verifiers (such as compilers or physical simulators) into an iterative self-correction loop has proven essential for reliable engineering synthesis~\cite{lai2025analogcoder}. Building upon these agentic principles, our work establishes a specialized closed-loop workflow for power electronics: we bridge theoretical analytical sizing with commercial distributor APIs for authentic component selection, and close the verification loop via automated NGSPICE transient simulations to guarantee functional schematic synthesis.

% ==========================================
% Section 3: Methodology / System Architecture
% ==========================================
\section{Proposed Automated Synthesis Workflow}
\label{sec:method}

\begin{figure*}[htbp]
  \centering
  % 请将下载好的架构图命名为 framework.png (或 .jpg) 并上传至 Overleaf 的 media/ 文件夹下
  \includegraphics[width=0.8\textwidth]{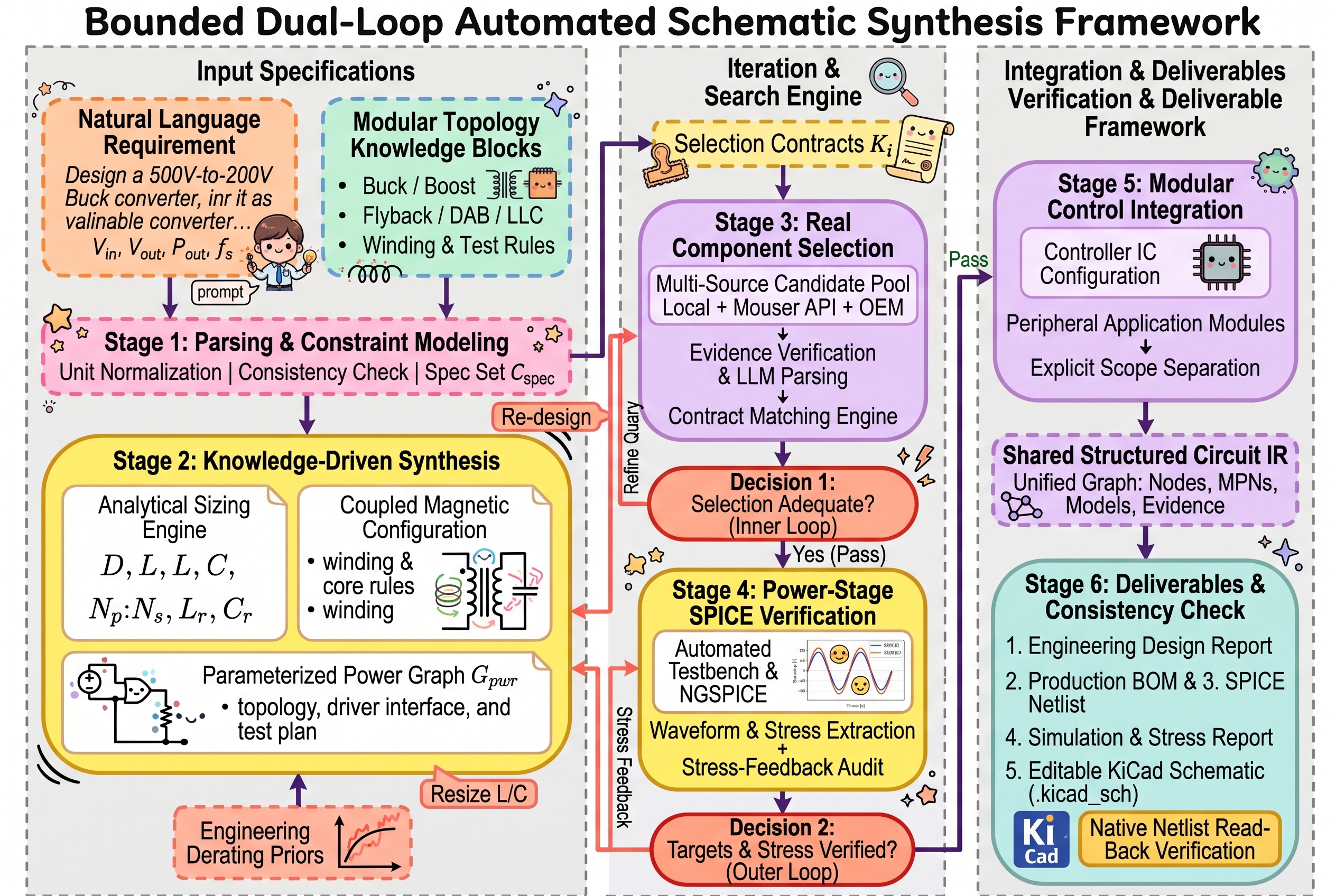}
  \caption{The proposed \textbf{Bounded Dual-Loop Automated Schematic Synthesis Framework}. The architecture is organized into three functional columns from left to right: (1) \textit{Input Specifications}, (2) \textit{Iteration \& Search Engine}, and (3) \textit{Integration \& Deliverables}. Deep purple solid arrows denote the forward synthesis and integration pipeline, whereas coral-orange solid arrows highlight the diagnostic feedback pathways (\texttt{Refine Query}, \texttt{Re-design}, \texttt{Stress Feedback}, and \texttt{Resize L/C}) governed by Decision Nodes 1 and 2.}
  \label{fig:architecture}
\end{figure*}

To translate unstructured natural language design requests into physically valid, manufacturing-ready power electronics hardware, we formulate the schematic design process as a knowledge-constrained, bounded dual-loop workflow. Rather than relying on open-loop text generation, our framework coordinates symbolic domain knowledge, multi-source commercial supply-chain evidence, and time-domain circuit simulation.

\subsection{System Overview and Bounded Dual-Loop Architecture}
\label{subsec:overview}

As illustrated in Figure~\ref{fig:architecture}, the overall \textbf{Bounded Dual-Loop Automated Schematic Synthesis Framework} is structured horizontally across three functional columns encompassing six interconnected stages:
\begin{enumerate}
    \item \textbf{Column 1 -- Input Specifications:} Encompasses \textit{Stage 1 (Parsing \& Constraint Modeling)} and \textit{Stage 2 (Knowledge-Driven Synthesis)}, bridging natural language requirements, modular topology knowledge blocks, and engineering derating priors into a parameterized power graph $\mathcal{G}_{\text{pwr}}$.
    \item \textbf{Column 2 -- Iteration \& Search Engine:} Encompasses \textit{Stage 3 (Real Component Selection)} and \textit{Stage 4 (Power-Stage SPICE Verification)}, executing contract-based device retrieval and automated NGSPICE verification governed by two diagnostic decision nodes.
    \item \textbf{Column 3 -- Integration \& Deliverables:} Encompasses \textit{Stage 5 (Modular Control Integration)} and \textit{Stage 6 (Deliverables \& Consistency Check)}, assembling peripheral control hardware and compiling multi-format deliverables via a Shared Structured Circuit Intermediate Representation (IR).
\end{enumerate}

In Figure~\ref{fig:architecture}, the primary forward progression is indicated by \textbf{deep purple solid arrows}, while diagnostic self-correction pathways are depicted by \textbf{coral-orange solid arrows}. Crucially, Column 2 operates as a \textbf{``Design--Selection--Simulation--Feedback'' bounded closed loop} governed by two diagnostic decision nodes:
\begin{itemize}
    \item \textbf{Inner Loop (\texttt{Decision 1: Selection Adequate?}):} Audits whether every physical component position in the synthesized topology is assigned a specific commercial part number backed by verified datasheet/distributor evidence and compliant electrical ratings. When unmet, the coral-orange feedback pathways either route back to Stage 3 (\texttt{Refine Query}) to expand candidate retrieval and evidence parsing, or escalate across the column boundary back to Stage 2 (\texttt{Re-design}) if physical COTS devices cannot satisfy the theoretical parameters. Upon passing (\texttt{Yes (Pass)}), execution proceeds to Stage 4.
    \item \textbf{Outer Loop (\texttt{Decision 2: Targets \& Stress Verified?}):} Audits whether the compiled circuit executes properly in NGSPICE and satisfies declared steady-state, ripple, and gate-drive targets, while comparing extracted time-domain electrical stresses against component contracts. If simulated stresses exceed initial estimates, the outer loop routes back to Stage 3 via the \texttt{Stress Feedback} pathway to update selection contracts and re-select higher-rated parts; if output voltage or ripple targets fail, it routes back to Stage 2 via the \texttt{Resize L/C} pathway to retune passive or modulation parameters. Once verified (\texttt{Pass}), the workflow advances to Column 3.
\end{itemize}

\paragraph{Bounded Termination and Execution Scope.} To guarantee deterministic engineering execution, user-defined functional requirements and acceptance boundaries remain immutable throughout the loops, while feedback iteratively adjusts candidate pools, allowed design degrees of freedom, and model bindings. The workflow terminates either when all verification targets are satisfied or when predefined computational, retrieval, and iteration budgets are exhausted—in which case the system outputs the best-effort design alongside explicit unresolved items and diagnostic logs. In our current implementation, the default basic execution path incorporates reference power-stage simulation with a single-pass stress-feedback re-selection, whereas a strict device-model-bound entry point supports multi-iteration bounded repair loops.

% ==========================================
% Stage 1 & Stage 2 (Column 1)
% ==========================================
\subsection{Input Specifications: Parsing and Knowledge-Driven Synthesis}
\label{subsec:synthesis}

\subsubsection{Stage 1: Parsing and Constraint Modeling}
As shown at the top of Column 1 in Figure~\ref{fig:architecture}, the workflow begins by establishing a formal interface between a \textbf{Natural Language Requirement} (e.g., \textit{``Design a 500V-to-200V Buck converter...''}) and electrical engineering constraints. Combining the user prompt with \textbf{Modular Topology Knowledge Blocks}, Stage 1 extracts key specification tuples—including topology family, input/output ratings ($V_{\text{in}}, V_{\text{out}}, P_{\text{out}}$), switching frequency $f_s$, ripple tolerances, and isolation requirements. The module performs automatic unit normalization, completeness checking, and consistency validation. Importantly, the parser distinguishes between mandatory user constraints, engineering preferences, and unspecified conditions, outputting a formal specification set $\mathcal{C}_{\text{spec}}$.

\subsubsection{Stage 2: Knowledge-Driven Synthesis}
Guided by $\mathcal{C}_{\text{spec}}$ and bottom-up \textbf{Engineering Derating Priors} (e.g., voltage and current safety margin curves), Stage 2 synthesizes the power stage through three coordinated sub-modules:
\begin{enumerate}
    \item \textbf{Analytical Sizing Engine:} Computes key circuit parameters—such as duty cycle $D$, filtering inductances and capacitances ($L, C$), transformer turns ratios ($N_p : N_s$), and resonant tank parameters ($L_r, C_r$)—across five supported topologies:
    \begin{itemize}
        \item \textit{Buck and Boost:} Passive components and duty cycles are sized via inductor volt-second balance and capacitor ampere-second balance under continuous conduction mode (CCM) ripple constraints:
        \begin{equation}
            L_{\text{buck}} = \frac{V_{\text{out}}(V_{\text{in}} - V_{\text{out}})}{\Delta I_L \cdot f_s \cdot V_{\text{in}}}, \quad C_{\text{out,buck}} = \frac{\Delta I_L}{8 \cdot f_s \cdot \Delta V_{\text{out}}}
            \label{eq:buck_boost}
        \end{equation}
        \item \textit{Flyback:} Magnetizing inductance $L_m$, turns ratio $N_p : N_s$, and operating points are determined from per-cycle inductive energy storage, volt-second demagnetization relations, and reflected voltage limits.
        \item \textit{Dual-Active Bridge (DAB):} Based on single-phase-shift (SPS) power transfer, the series inductance $L_s$ and reference phase-shift ratio $\phi$ are synthesized via:
        \begin{equation}
            P_{\text{DAB}} = \frac{n V_{\text{in}} V_{\text{out}}}{2 f_s L_s} \phi (1 - |\phi|)
            \label{eq:dab}
        \end{equation}
        \item \textit{Half-Bridge LLC:} Utilizing First Harmonic Approximation (FHA), resonant parameters ($L_r, C_r, L_m$) are synthesized to ensure inductive-region operation for Zero-Voltage Switching (ZVS).
    \end{itemize}
    \item \textbf{Coupled Magnetic Configuration:} For transformer-isolated topologies (Flyback, DAB, LLC), literature-backed winding configurations and physical core rules are incorporated so that realistic magnetic parameters directly participate in circuit synthesis.
    \item \textbf{Parameterized Power Graph ($\mathcal{G}_{\text{pwr}}$):} Aggregates the synthesized topology connectivity, gate-driver interface requirements, and an executable verification test plan, forwarding them to Column 2.
\end{enumerate}

% ==========================================
% Stage 3 & Stage 4 (Column 2)
% ==========================================
\subsection{Iteration and Search Engine: Selection and SPICE Verification}
\label{subsec:selection_sim}

\subsubsection{Stage 3: Real Component Selection}
Entering Column 2 of Figure~\ref{fig:architecture}, the requirements embedded in $\mathcal{G}_{\text{pwr}}$ are formalized into position-specific \textbf{Selection Contracts ($\mathcal{K}_i$)}, defining nominal value bounds, voltage/current/power ratings, supply voltages, and interface rules. Stage 3 executes three sequential steps:
\begin{enumerate}
    \item \textbf{Multi-Source Candidate Pool:} Generates parametric search queries to retrieve candidate parts from local curated catalogs, the real-time \textbf{Mouser API}, and OEM datasheets.
    \item \textbf{Evidence Verification \& LLM Parsing:} Performs identity association, unit normalization, and provenance auditing. For unstructured datasheet text requiring semantic interpretation, the LLM proposes interpretations that are strictly cross-checked against original document excerpts and deterministic rules.
    \item \textbf{Contract Matching Engine:} Evaluates normalized candidates item-by-item against $\mathcal{K}_i$ while accounting for package type, thermal limits, operating frequency, and parameter guarantee types.
\end{enumerate}

As summarized in the upper half of Table~\ref{tab:routing}, \textbf{Decision 1} governs the inner loop: if selection is inadequate, the system invokes either the \texttt{Refine Query} pathway back to Stage 3 or the \texttt{Re-design} pathway back to Stage 2.

\begin{table}[htbp]
 \caption{Diagnostic Routing Logic and Solid Feedback Pathways at Decision Nodes 1 and 2}
  \centering
  \small
  \begin{tabular}{p{3.0cm}p{4.0cm}p{3.4cm}p{4.8cm}}
    \toprule
    \textbf{Decision Node} & \textbf{Failure / Unmet Condition} & \textbf{Return Target (Figure~\ref{fig:architecture} Label)} & \textbf{Corrective Action Taken} \\
    \midrule
    \textbf{Decision 1: Selection Adequate?} \newline \textit{(Inner Loop)} 
    & Insufficient candidates or category mismatch & Stage 3: Candidate Pool \newline (\texttt{Refine Query}) & Adjust search keywords; expand manufacturer scope or distributor API sources. \\
    \cmidrule(l){2-4}
    & Missing critical parameters in candidate records & Stage 3: Evidence Verification \newline (\texttt{Refine Query}) & Supplement datasheets; extract missing parameters; verify part identity and conditions. \\
    \cmidrule(l){2-4}
    & Candidate key ratings violate contract $\mathcal{K}_i$ & Stage 3: Contract Matching \newline (\texttt{Refine Query}) & Search for alternative commercial models satisfying electrical/thermal constraints. \\
    \cmidrule(l){2-4}
    & Design parameters unachievable by real COTS parts & Stage 2: Circuit Synthesis \newline (\texttt{Re-design}) & Adjust permissible design parameters or physical configurations while keeping user specs fixed. \\
    \midrule
    \textbf{Decision 2: Targets \& Stress Verified?} \newline \textit{(Outer Loop)} 
    & Output voltage or ripple violates target tolerances & Stage 2: Circuit Synthesis \newline (\texttt{Resize L/C}) & Retune passive $L/C$ values, reference modulation parameters, or structural configurations. \\
    \cmidrule(l){2-4}
    & Simulated stress elevates device rating requirements & Stage 3: Component Selection \newline (\texttt{Stress Feedback}) & Update contract $\mathcal{K}_i$ at affected positions with simulated stresses; trigger targeted re-selection. \\
    \cmidrule(l){2-4}
    & Model binding, port, or numerical solver issue & Stage 4: Testbench \& NGSPICE \newline (\textit{Internal Sim Repair}) & Audit subcircuit bindings, pin interfaces, initial conditions, and numerical solver settings. \\
    \cmidrule(l){2-4}
    & Incomplete operating condition or measurement coverage & Stage 4: Test Plan Execution \newline (\textit{Internal Sim Repair}) & Supplement missing operating sweeps or measurements before granting a \texttt{Pass}. \\
    \bottomrule
  \end{tabular}
  \label{tab:routing}
\end{table}

\subsubsection{Stage 4: Power-Stage SPICE Verification}
Once Decision 1 yields \texttt{Yes (Pass)}, Stage 4 compiles the circuit into an executable SPICE netlist and invokes an \textbf{Automated Testbench \& NGSPICE} engine (supplying input sources, loads, auxiliary supplies, and gate-drive modulation signals) to run nominal and fine-step transient simulations. Next, the \textbf{Waveform \& Stress Extraction} and \textbf{Stress-Feedback Audit} modules evaluate steady-state output voltage, voltage/current ripples, and gate-drive behavior, while extracting terminal peak voltages, peak currents, and RMS currents across all components.

As detailed in the lower half of Table~\ref{tab:routing}, \textbf{Decision 2} governs the outer loop: electrical target violations trigger the \texttt{Resize L/C} pathway back to Stage 2, whereas elevated component stresses trigger the \texttt{Stress Feedback} pathway back to Stage 3 to update $\mathcal{K}_i$ and re-select affected devices.

% ==========================================
% Stage 5 & Stage 6 (Column 3)
% ==========================================
\subsection{Integration and Deliverables: Control Hardware and Consistency Check}
\label{subsec:deliverables}

\subsubsection{Stage 5: Modular Control Integration}
Upon passing Decision 2 (\texttt{Pass}), execution transitions to Column 3 of Figure~\ref{fig:architecture}. Stage 5 performs \textbf{Controller IC Configuration} (supporting analog PWM, digital PWM, and integrated half-bridge controllers) and assembles standardized \textbf{Peripheral Application Modules}—including supply decoupling, voltage references, timing networks, reset circuits, sensing interfaces, logic level shifting, and gate-driver connections. Through \textbf{Explicit Scope Separation}, Stage 5 focuses strictly on control hardware selection and pin-level connectivity (excluding control algorithm synthesis, DSP firmware, or closed-loop stability design) and stores the control hardware results separately from the verified power-stage simulation model.

\subsubsection{Stage 6: Deliverables and Consistency Check}
Both the verified power-stage design and modular control hardware are unified into a \textbf{Shared Structured Circuit IR}—a central graph linking topological nodes, verified MPNs, simulation models, and datasheet evidence. From this unified IR, Stage 6 generates five synchronized engineering deliverables:
\begin{enumerate}
    \item \textbf{Engineering Design Report:} Summarizing parsed requirements, analytical sizing derivations, selection evidence, and design boundaries.
    \item \textbf{Production BOM} and \textbf{3. SPICE Netlist:} Providing verified MPNs, manufacturers, quantities, key ratings, and executable SPICE netlists.
    \setcounter{enumi}{3}
    \item \textbf{Simulation \& Stress Report:} Archiving operating conditions, measured metrics, voltage/current waveforms, stress profiles, and model hierarchy declarations.
    \item \textbf{Editable KiCad Schematic (\texttt{.kicad\_sch}):} Generates an editable schematic file coupled with an automated \textbf{Native Netlist Read-Back Verification} module, which exports the native netlist back from KiCad to verify 100\% topological consistency against the Shared Structured Circuit IR.
\end{enumerate}

% ==========================================
% Section 4: Experimental Setup & Case Studies
% ==========================================
\section{Case Studies and Experimental Evaluation}
\label{sec:experiments}

To evaluate the effectiveness, topological generalizability, and engineering rigor of the proposed workflow, we first detail our system implementation, then present two comprehensive end-to-end synthesis case studies—a non-isolated Buck converter and an isolated Dual-Active Bridge (DAB) converter—and finally conduct a quantitative and diagnostic comparison against a one-shot Vanilla LLM baseline.

\subsection{Implementation Details}
\label{subsec:impl}

The proposed framework is implemented in modular Python, orchestrating requirement compilation, topology knowledge synthesis, evidence-backed component selection, SPICE simulation verification, and multi-format deliverable generation. All experiments were executed locally on a 64-bit Windows 10 workstation equipped with an Intel Core i7-13700F CPU (16 cores, 24 threads) and 32~GB of RAM. The primary workflow engine runs under Python 3.14.5 (with NumPy 2.4.6 and Matplotlib 3.10.9 for waveform post-processing), while a dedicated Python 3.12.14 runtime handles PDF datasheet parsing and page rendering via \texttt{pdfplumber} and \texttt{pypdfium2}. Time-domain circuit simulations are solved using \texttt{ngspice 47}, and editable schematics along with native netlist round-trip checks are generated via \texttt{KiCad CLI 10.0.0}.

\begin{table}[htbp]
 \caption{Summary of Core Implementation and Experimental Settings}
  \centering
  \small
  \begin{tabular}{ll}
    \toprule
    \textbf{Category} & \textbf{Configuration / Setting} \\
    \midrule
    Default Workflow Engine & Deterministic parsing \& topology sizing + basic real-part selection \\
    Default Simulation Scope & Externally modulated power stage and gate driver (\texttt{ucc27714} adapter) \\
    Basic Stress-Feedback Loop & Single-pass simulation-stress-informed component re-selection \\
    Selection Tolerance Windows & Resistors: $\pm 10\%$; Inductors \& Capacitors: $\pm 25\%$ \\
    Baseline Derating Rules & Voltage: $\ge 1.2\times V_{\text{peak}}$; Current (MOSFET/Ind./Diode): $\ge 1.25\times I_{\text{rms/avg}}$ \\
    Circuit Simulator & \texttt{ngspice 47} ($T = 27^\circ\text{C}$, $\text{reltol}=10^{-4}$, $\text{abstol}=1\,\mu\text{A}$, $\text{vntol}=1\,\mu\text{V}$) \\
    Schematic \& Verification Tool & \texttt{KiCad CLI 10.0.0} (S-expression export \& native netlist round-trip check) \\
    Semantic Datasheet LLM & \texttt{qwen3.5-plus-2026-04-20} (Alibaba Cloud API, no local fine-tuning) \\
    LLM Output \& Context Budget & \texttt{json\_object} mode, \texttt{enable\_thinking=false}, $\le 6000$ tokens, $\le 4$ pages/task \\
    \bottomrule
  \end{tabular}
  \label{tab:impl_settings}
\end{table}

Table~\ref{tab:impl_settings} summarizes the key system configurations. In our default power-stage execution path (\texttt{language\_spice / power\_stage / basic}), requirement parsing, topology equations, and baseline contract matching are executed deterministically, followed by externally modulated power-stage and gate-driver simulation in \texttt{ngspice 47} with one bounded stress-feedback re-selection pass. For semantic datasheet extraction tasks, the system invokes \texttt{qwen3.5-plus-2026-04-20} via a hosted Chat Completions endpoint (client-side sampling parameters such as temperature and top-$p$ are left unspecified at their server defaults). LLM outputs are treated strictly as candidate proposals that must pass deterministic provenance and document grounding checks against source PDFs before adoption. Importantly, reference power-stage simulation models, commercial BOM realizations, and modular controller-hardware proposals are logged separately to maintain transparent verification boundaries.

% ==========================================
% 4.2 Case Study 1: Buck Converter
% ==========================================
\subsection{End-to-End Case Study I: Non-Isolated Buck Converter}
\label{subsec:case_buck}

\paragraph{Design Task and Knowledge-Driven Synthesis.} 
The first case study evaluates a non-isolated step-down converter triggered by the natural language prompt: \textit{``Buck input 24-36V output 12V 2A 80kHz non-isolated ripple 5\%''}. The requirement parser normalizes the specifications into $V_{\text{in}} \in [24, 36]\,\text{V}$, $V_{\text{out}} = 12\,\text{V}$, $I_{\text{out}} = 2\,\text{A}$ ($P_{\text{out}} = 24\,\text{W}$), $f_s = 80\,\text{kHz}$, and a peak-to-peak output voltage ripple limit of $\Delta V_{\text{out,max}} = 0.6\,\text{V}_{\text{pp}}$ ($5\%$). Invoking the Buck knowledge module under volt-second balance and ripple constraints yields theoretical reference parameters of $L_{\text{main}} = 166.667\,\mu\text{H}$, $C_{\text{out}} = 151.982\,\mu\text{F}$, $C_{\text{in}} = 30.396\,\mu\text{F}$, current-sense resistor $R_{\text{CS}} = 3.846\,\text{m}\Omega$, gate resistor $R_{\text{gate}} = 10\,\Omega$, and bootstrap capacitor $C_{\text{boot}} = 1.5\,\mu\text{F}$.

\begin{table}[htbp]
 \caption{Synthesized Parameters and Selected Core Commercial Components for Case Study I (Buck) and Case Study II (DAB)}
  \centering
  \footnotesize
  \setlength{\tabcolsep}{4.5pt}
  \resizebox{\textwidth}{!}{%
  \begin{tabular}{lllll}
    \toprule
    \textbf{Functional Role (Designator)} & \textbf{Theoretical / Ref.} & \textbf{Selected Nominal} & \textbf{Verified Part Number (Mfr.)} & \textbf{Qty} \\
    \midrule
    \multicolumn{5}{l}{\textbf{\textit{Case Study I: Non-Isolated Buck Converter (29 physical positions, 22 unique MPNs in total)}}} \\
    \midrule
    High-Side Switch (\texttt{QSW / Q1}) & $V_{\text{in,max}}=36\,\text{V}$ & $600\,\text{V}, 41\,\text{m}\Omega$ & \texttt{IPW60R041P6FKSA1} (Infineon) & 1 \\
    Freewheeling Diode (\texttt{DFREE / D1}) & $I_{\text{out}}=2\,\text{A}$ & $200\,\text{V}, 3\,\text{A}$ & \texttt{UF3004-HF} (Comchip) & 1 \\
    Main Inductor (\texttt{LMAIN / L1}) & $166.67\,\mu\text{H}$ & $180\,\mu\text{H}$ & \texttt{760800301} (W{\"u}rth Elektronik) & 1 \\
    Output Capacitor (\texttt{COUT / C1}) & $151.98\,\mu\text{F}$ & $150\,\mu\text{F}$ ($200\,\text{V}$) & \texttt{EKXJ201ELL151MJ50S} (Chemi-Con) & 1 \\
    Input Capacitor (\texttt{CIN / C2}) & $30.40\,\mu\text{F}$ & $30\,\mu\text{F}$ ($600\,\text{V}$) & \texttt{C44AHFP5300ZC0J} (KEMET) & 1 \\
    Sense Resistor (\texttt{RCS / R1}) & $3.85\,\text{m}\Omega$ & $3.9\,\text{m}\Omega$ & \texttt{SL1TTE3L9J} (KOA Speer) & 1 \\
    Gate Driver (\texttt{GD\_main / U1}) & High/Low-Side & Half-Bridge Driver & \texttt{UCC27714D} (Texas Instruments) & 1 \\
    Analog Controller (\texttt{CONTROL / U3}) & Peripheral Mod. & Current-Mode PWM & \texttt{UCC28C53DR} (Texas Instruments) & 1 \\
    \midrule
    \multicolumn{5}{l}{\textbf{\textit{Case Study II: Isolated Dual-Active Bridge (DAB) Converter (87 physical positions, 20 unique MPNs in total)}}} \\
    \midrule
    Full-Bridge Switches (\texttt{Q1--Q8}) & $V_{\text{in}}=48\,\text{V}$ & $600\,\text{V}, 41\,\text{m}\Omega$ & \texttt{IPW60R041P6FKSA1} (Infineon) & 8 \\
    Series Inductor (\texttt{LSER / L1}) & $101.38\,\mu\text{H}$ & $100\,\mu\text{H}$ & \texttt{ATCA-01-101M-V} (ABRACON) & 1 \\
    Isolation Transformer (\texttt{TMAIN / T1}) & $N_s/N_p=0.5, L_m=420.8\,\mu\text{H}$ & 4P:2S Winding Config. & \texttt{HPH6-0325L} (Coilcraft) & 1 \\
    In/Out Capacitors (\texttt{CIN, COUT}) & $20.83\,\mu\text{F}$ & $22\,\mu\text{F}$ ($450\,\text{V}$) & \texttt{B32774D4226K} (TDK) & 2 \\
    Gate Drivers (\texttt{GD\_Q1--GD\_Q8}) & 8-Ch Drive & Half-Bridge Driver & \texttt{UCC27714D} (Texas Instruments) & 8 \\
    Digital Controller (\texttt{CONTROL / U13}) & SPS PWM Gen. & C2000 32-bit MCU & \texttt{F280039CSPM} (Texas Instruments) & 1 \\
    Digital Isolators (\texttt{U14, U15}) & Iso. PWM Interface & Quad-Ch Isolator & \texttt{ISO7740FDWR} (Texas Instruments) & 2 \\
    \bottomrule
  \end{tabular}%
  }
  \label{tab:case_bom}
\end{table}

\paragraph{Component Selection and Control Hardware Integration.} 
Within the declared nominal tolerance windows, the selection engine maps theoretical demands to commercial catalog parts and audits them against simulated reference stresses. The base power-stage procurement list comprises 14 component positions; after integrating the modular analog PWM controller (\texttt{UCC28C53DR}) and its application peripherals (supply/reference decoupling, $RT/CT$ timing, feedback divider, current-sense RC filter, and compensation network options), the complete hardware proposal encompasses \textbf{29 physical positions across 22 unique MPNs}, with 100\% part-number assignment. The upper half of Table~\ref{tab:case_bom} lists the core selected devices.

\begin{table}[htbp]
 \caption{NGSPICE Reference Verification Results Across Operating Conditions for Case Study I (Buck) and Case Study II (DAB)}
  \centering
  \footnotesize
  \setlength{\tabcolsep}{5pt}
  \resizebox{\textwidth}{!}{%
  \begin{tabular}{lllcccc}
    \toprule
    \textbf{Case (Eval. Window)} & \textbf{Test Condition} & \textbf{Operating Point} & \textbf{Max Step} & \textbf{Mean $V_{\text{out}}$ (V)} & \textbf{Mean $I_{\text{out}}$ (A)} & \textbf{Ripple $V_{\text{pp}}$ (V) / Status} \\
    \midrule
    \textbf{Buck} & \texttt{vin\_low} & $24\,\text{V} / 6\,\Omega$ & $312.5\,\text{ns}$ & $11.8639$ & $1.9773$ & $0.0671$ (\textbf{PASS}) \\
    \textit{(6.4--8.0 ms)} & \texttt{vin\_high} & $36\,\text{V} / 6\,\Omega$ & $312.5\,\text{ns}$ & $11.8363$ & $1.9727$ & $0.0693$ (\textbf{PASS}) \\
    & \texttt{half\_load} & $24\,\text{V} / 12\,\Omega$ & $312.5\,\text{ns}$ & $11.8926$ & $0.9910$ & $0.1399$ (\textbf{PASS}) \\
    & \texttt{fine\_timestep} & $24\,\text{V} / 6\,\Omega$ & $156.25\,\text{ns}$ & $11.8639$ & $1.9773$ & $0.0670$ (\textbf{PASS}) \\
    \midrule
    \textbf{DAB} & \texttt{nominal} & $48\,\text{V} / 48\,\Omega$ ($\phi=0.06$) & $166.7\,\text{ns}$ & $22.7423$ & $0.4738$ & $0.1788$ (\textbf{PASS}) \\
    \textit{(4.8--6.0 ms)} & \texttt{fine\_timestep} & $48\,\text{V} / 48\,\Omega$ ($\phi=0.06$) & $83.3\,\text{ns}$ & $22.7422$ & $0.4738$ & $0.1795$ (\textbf{PASS}) \\
    \bottomrule
  \end{tabular}%
  }
  \label{tab:case_sim}
\end{table}

\begin{figure}[htbp]
  \centering
  \begin{minipage}[b]{0.49\textwidth}
    \centering
    \includegraphics[width=\textwidth]{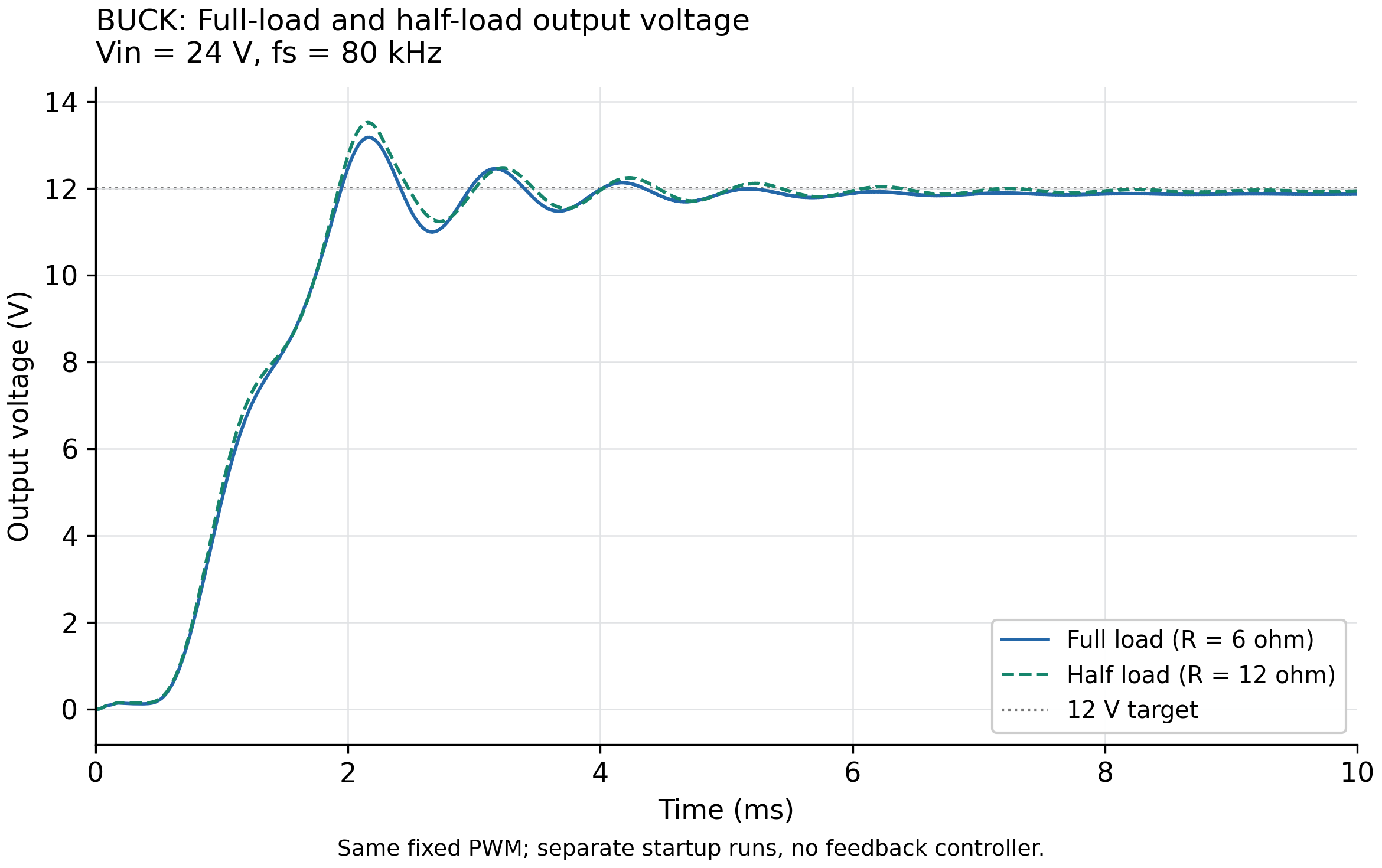}
    \vspace{2pt}
    {\footnotesize (a) Non-isolated Buck converter ($V_{\text{in}}=24\,\text{V}, f_s=80\,\text{kHz}$)}
  \end{minipage}
  \hfill
  \begin{minipage}[b]{0.49\textwidth}
    \centering
    \includegraphics[width=\textwidth]{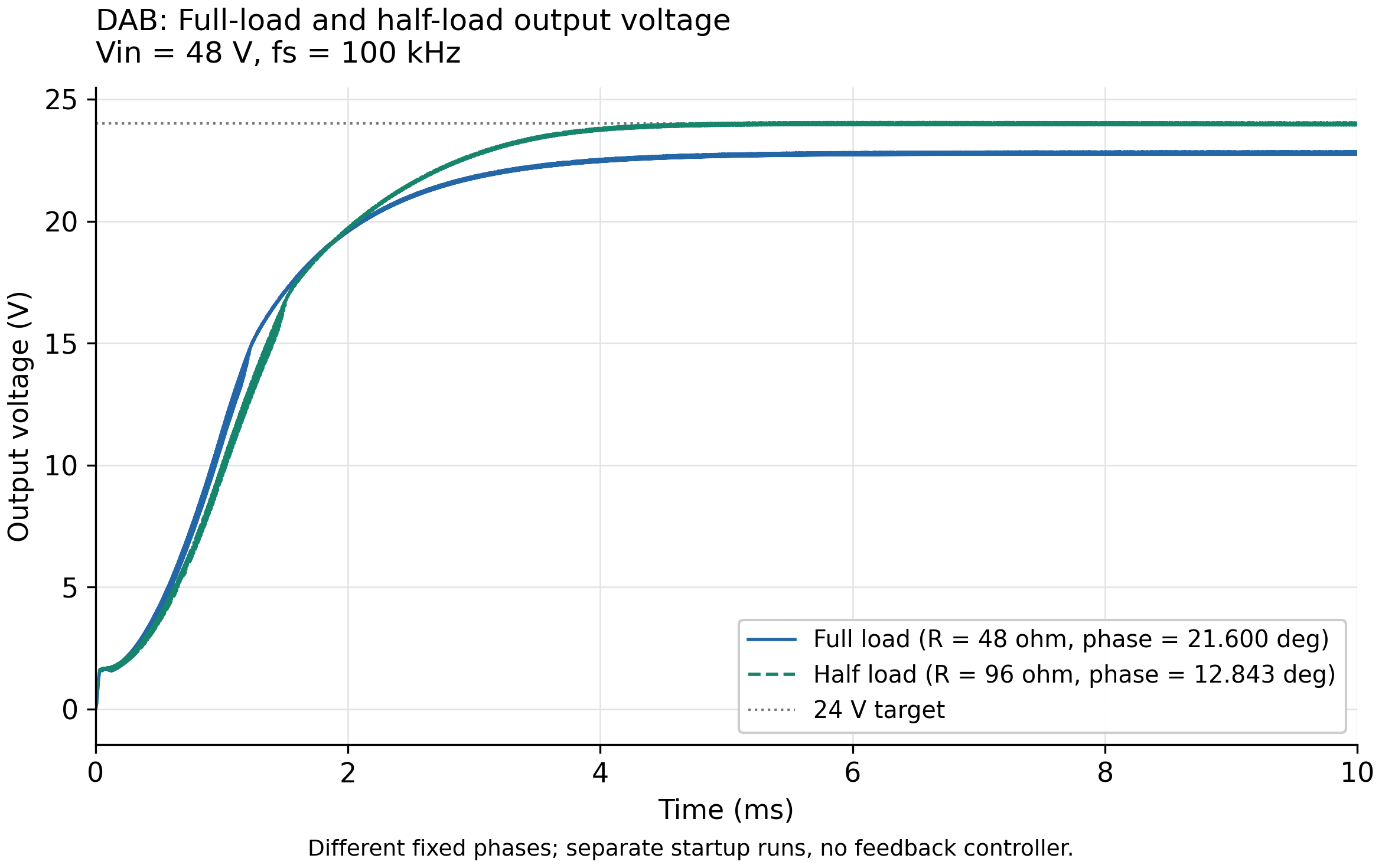}
    \vspace{2pt}
    {\footnotesize (b) Isolated DAB converter ($V_{\text{in}}=48\,\text{V}, f_s=100\,\text{kHz}$)}
  \end{minipage}
  \caption{Simulated \texttt{ngspice} startup and steady-state output voltage ($V_{\text{out}}$) waveforms under full-load and half-load operating conditions without closed-loop feedback controllers: (a) Buck converter operating under a fixed PWM duty cycle at full load ($R = 6\,\Omega$) and half load ($R = 12\,\Omega$) against the $12\,\text{V}$ target; and (b) DAB converter operating under topology-synthesized single-phase-shift (SPS) angles at full load ($R = 48\,\Omega$, phase $= 21.600^\circ$) and half load ($R = 96\,\Omega$, phase $= 12.843^\circ$) against the $24\,\text{V}$ target.}
  \label{fig:waveforms}
\end{figure}

\paragraph{NGSPICE Verification and Deliverable Consistency.} 
The power-stage and gate-driver reference circuit (utilizing the source-derived \texttt{ti\_ucc27714\_enable\_port} model) was verified across four operating conditions over an $8\,\text{ms}$ horizon (evaluated in the final $20\%$ window, $6.4\text{--}8.0\,\text{ms}$). As shown in the upper half of Table~\ref{tab:case_sim} and Figure~\ref{fig:waveforms}(a), all four test cases—low input ($24\,\text{V}$), high input ($36\,\text{V}$), half load ($12\,\Omega$), and halved timestep ($156.25\,\text{ns}$)—passed all verification criteria. Specifically, under \texttt{vin\_low}, the converter achieved a mean output voltage of $11.8639\,\text{V}$ and a peak-to-peak ripple of $0.0671\,\text{V}_{\text{pp}}$ (well below the $0.6\,\text{V}_{\text{pp}}$ limit). Numerical sensitivity checks between nominal and fine timesteps yielded a negligible mean voltage discrepancy of $0.0021\,\text{mV}$ and ripple discrepancy of $0.0498\,\text{mV}$ (against a $60\,\text{mV}$ threshold). Finally, the generated KiCad schematic passed native netlist round-trip verification (\texttt{NATIVE\_ROUNDTRIP\_CONNECTIVITY\_VERIFIED}) with \textbf{83 out of 83 connected terminals} matching the internal circuit representation exactly.

% ==========================================
% 4.3 Case Study 2: Isolated DAB Converter
% ==========================================
\subsection{End-to-End Case Study II: Isolated Dual-Active Bridge (DAB) Converter}
\label{subsec:case_dab}

\paragraph{Design Task and Coupled Magnetic Synthesis.} 
To demonstrate scalability to complex isolated bridge topologies, we prompted the system with: \textit{``DAB input 48V output 24V 0.5A 100kHz isolated ripple 5\%''} ($P_{\text{out}} = 12\,\text{W}$, ripple limit $\Delta V_{\text{out,max}} = 1.2\,\text{V}_{\text{pp}}$). The DAB knowledge module applies Single-Phase-Shift (SPS) power transfer analysis, synthesizing a nominal phase-shift ratio of $\phi = 0.06$ cycles ($0.6\,\mu\text{s}$ or $21.6^\circ$ at $100\,\text{kHz}$, with $50\%$ duty cycle and $150\,\text{ns}$ deadtime), a required series energy-transfer inductance of $L_{\text{ser}} = 101.376\,\mu\text{H}$, and filtering capacitances of $C_{\text{in}} = C_{\text{out}} = 20.833\,\mu\text{F}$. 

Crucially, rather than assuming an idealized textbook transformer, the synthesis module selects a commercial multi-winding transformer (\texttt{Coilcraft HPH6-0325L}) and synthesizes a concrete physical winding configuration: four windings are connected in series on the primary side (terminals $1\text{--}9$, bridging $12\text{--}2$, $11\text{--}3$, and $10\text{--}4$) and two windings in series on the secondary side (terminals $5\text{--}7$, bridging $8\text{--}6$). This physical configuration realizes the exact required secondary-to-primary turns ratio of $N_s / N_p = 0.5$ and a primary magnetizing inductance of $L_m = 420.8\,\mu\text{H}$.

\paragraph{Component Selection, Simulation, and KiCad Verification.} 
As summarized in the lower halves of Tables~\ref{tab:case_bom} and~\ref{tab:case_sim}, the base power-stage BOM contains 57 positions, which expands to \textbf{87 enumerated physical positions across 20 unique MPNs} after integrating the digital control hardware—featuring a Texas Instruments C2000 MCU (\texttt{F280039CSPM}) with segregated $3.3\,\text{V}$ I/O and core supply decoupling, ADC sensing interfaces, and two quad-channel digital isolators (\texttt{ISO7740FDWR}) routing eight PWM channels across the galvanic isolation barrier to eight \texttt{UCC27714D} gate drivers. In \texttt{ngspice 47} verification (Figure~\ref{fig:waveforms}(b)), the nominal operating point achieved a mean output voltage of $22.7423\,\text{V}$ ($-5.24\%$ deviation from $24\,\text{V}$, well within the $\pm 10\%$ open-loop reference tolerance band) and a peak-to-peak ripple of $0.1788\,\text{V}_{\text{pp}}$ ($0.745\%$, far below the $5\%$ limit). Halving the maximum timestep to $83.3\,\text{ns}$ yielded virtually identical results ($\Delta V_{\text{mean}} = 0.0765\,\text{mV}$). Native KiCad netlist round-trip verification confirmed 100\% connectivity fidelity across all \textbf{341 out of 341 connected terminals}.

% ==========================================
% 4.4 Comparative Evaluation: Proposed vs. Vanilla LLM
% ==========================================
\subsection{Comparative Evaluation: Proposed Framework vs. Vanilla LLM}
\label{subsec:comparison}

\paragraph{Experimental Protocol.} 
To rigorously quantify the value of integrating domain knowledge, evidence verification, and simulation feedback, we compared our complete workflow against a one-shot, tool-free \textbf{Vanilla LLM} baseline (\texttt{qwen3.5-plus-2026-04-20} with $\text{temperature}=0$, $\text{top\_p}=1$, $\text{max\_tokens}=10000$, and \texttt{enable\_thinking=false}) on the exact same Buck and DAB design tasks. Both methods were evaluated under identical \texttt{ngspice 47} testbenches and audited against source distributor/OEM documentation across eight multi-dimensional engineering metrics.

\begin{table}[htbp]
 \caption{Quantitative Comparison Between the Proposed Complete Workflow and the Vanilla LLM Baseline Across the Frozen Buck and DAB Benchmark Tasks}
  \centering
  \small
  \setlength{\tabcolsep}{8pt}
  \begin{tabular}{p{8.2cm}cc}
    \toprule
    \textbf{Evaluation Metric} & \textbf{Proposed Framework} & \textbf{Vanilla LLM (One-Shot)} \\
    \midrule
    Verified Part Positions (All Enumerated Positions) & \textbf{98.6\%} (70 / 71) & 22.2\% (4 / 18) \\
    Verified Power-Core Component Positions$^{\dagger}$ & \textbf{94.4\%} (17 / 18) & 15.4\% (2 / 13) \\
    Verified Unique Part Numbers (Deduplicated MPNs) & \textbf{100.0\%} (17 / 17) & 21.4\% (3 / 14) \\
    Source Nominal Value Consistency with Design Target & \textbf{97.7\%} (43 / 44) & 0.0\% (0 / 7) \\
    Reference Parameter Evidence Coverage (Audited Fields) & \textbf{97.0\%} (162 / 167) & 7.5\% (4 / 53) \\
    Switch Control Reference Consistency (Gate-Return to Source) & \textbf{100.0\%} (9 / 9) & 60.0\% (3 / 5) \\
    Non-Bypass Power-Path Connectivity (In/Out Positive Nodes) & \textbf{100.0\%} (2 / 2) & 0.0\% (0 / 2) \\
    Executable \& Convergent SPICE Simulation & \textbf{100.0\%} (2 / 2) & 0.0\% (0 / 2) \\
    Electrical Specification Pass Rate ($V_{\text{out}}, I_{\text{out}}$ \& Ripple) & \textbf{100.0\%} (2 / 2) & 0.0\% (0 / 2) \\
    \bottomrule
    \multicolumn{3}{p{15.2cm}}{\footnotesize $^{\dagger}$Excludes gate drivers, driver peripheral passives (\texttt{GD\_*}), and controllers to prevent repetitive auxiliary positions from dominating the ratio.}
  \end{tabular}
  \label{tab:comparison_main}
\end{table}

\paragraph{Quantitative Results and Diagnostic Analysis.} 
As demonstrated in Table~\ref{tab:comparison_main}, the proposed framework substantially outperforms the Vanilla LLM across every verification dimension:
\begin{enumerate}
    \item \textbf{Elimination of Supply-Chain Hallucinations and Unverified Placeholders:} In the warm-library functional evaluation, our framework achieved a \textbf{98.6\% (70/71)} position verification rate and \textbf{100\% (17/17)} deduplicated MPN verification rate (with only a single Buck inductor position left unverified and transparently flagged as \texttt{UNKNOWN} rather than fabricated), covering \textbf{97.0\% (162/167)} of required datasheet parameter fields. Conversely, only \textbf{22.2\% (4/18)} of Vanilla LLM component positions could be verified against source evidence. For the DAB magnetic components, the Vanilla LLM failed to select real commercial parts, outputting generic text placeholders (\texttt{DAB\_LEAKAGE\_IND} and \texttt{DAB\_XFMR\_2\_1}), whereas our framework selected verified COTS magnetics (\texttt{ATCA-01-101M-V} and \texttt{HPH6-0325L}) with explicit physical pin bindings.
    \item \textbf{Prevention of Severe Nominal Mismatches and Voltage Rating Violations:} Auditing selected part numbers against official datasheets revealed a \textbf{97.7\% (43/44)} nominal value match for our framework versus \textbf{0.0\% (0/7)} for the Vanilla LLM. Notably, in the Buck task, the Vanilla LLM assigned resistor \texttt{RC0402FR-0710RL} for an intended $10\,\text{m}\Omega$ ($0.01\,\Omega$) capacitor ESR, whereas the actual datasheet resistance of that part number is $10\,\Omega$—a \textbf{$1000\times$ magnitude error}. Even more critically, as shown in Table~\ref{tab:failure_cases}, the Vanilla LLM committed severe self-contradictory voltage rating violations: in the $36\,\text{V}$-input, $12\,\text{V}$-output Buck converter, it selected input and output capacitors (\texttt{T491A476K010AT} and \texttt{GRM32EC71A107KE15L}) rated at only \textbf{$10\,\text{V}$}, which would cause immediate dielectric breakdown in hardware.
    \item \textbf{Controller Role Mismatch:} While our framework paired appropriate controllers (\texttt{UCC28C53DR} and \texttt{F280039CSPM}) with verified external gate drivers (\texttt{UCC27714D}), the Vanilla LLM selected \texttt{TPS54331DDCR} as an ``$80\,\text{kHz}$ timing reference'' to drive an external MOSFET in the $36\,\text{V}$ Buck circuit. However, according to Texas Instruments documentation, the \texttt{TPS54331} family is an integrated-switch buck converter with a maximum input rating of $28\,\text{V}$ and a fixed $570\,\text{kHz}$ switching frequency—rendering it fundamentally incompatible with both the $36\,\text{V}$ bus and the $80\,\text{kHz}$ external-switch architecture.
    \item \textbf{Topological Connectivity and Simulation Convergence:} Neither of the circuits generated by the Vanilla LLM could execute in \texttt{ngspice} (\textbf{0\% convergence}). Structural audits revealed that in the Vanilla Buck netlist, the positive input terminal was connected solely to the input shunt capacitor without any series power path to the switch, while the PWM voltage source and driver output simultaneously drove node \texttt{GATE\_SIG}, causing a singular matrix and ``timestep too small'' abort. Similarly, the Vanilla DAB netlist left the positive output terminal isolated from the power transfer stage and misbound the stimulus port name. In contrast, our framework achieved \textbf{100\% structural connectivity and simulation pass rates} across all operating conditions (achieving $V_{\text{out}} = 11.8839\,\text{V}, V_{\text{pp}} = 0.0820\,\text{V}$ on Buck \texttt{vin\_low} and $V_{\text{out}} = 23.2966\,\text{V}, V_{\text{pp}} = 0.1562\,\text{V}$ on DAB \texttt{nominal} under the unified functional evaluation benchmark).
\end{enumerate}

\begin{table}[htbp]
 \caption{Diagnostic Breakdown of Representative Component and Structural Failures in Vanilla LLM vs. Proposed Framework}
  \centering
  \footnotesize
  \setlength{\tabcolsep}{4.5pt}
  \begin{tabular}{p{2.6cm}p{2.2cm}p{5.2cm}p{5.0cm}}
    \toprule
    \textbf{Audit Dimension} & \textbf{Case \& Pos.} & \textbf{Vanilla LLM Output \& Failure Mode} & \textbf{Proposed Framework Output \& Verification} \\
    \midrule
    \textbf{Nominal Value Match} 
    & Buck \texttt{R\_ESR} vs. \texttt{RCS} 
    & \texttt{RC0402FR-0710RL}: Target $10\,\text{m}\Omega$, actual datasheet value is $10\,\Omega$ (\textbf{$1000\times$ error}, \texttt{FAIL}). 
    & \texttt{SL1TTE3L9J}: Target $3.85\,\text{m}\Omega$, actual part is $3.9\,\text{m}\Omega$ (\textbf{Within $\pm 10\%$ window}, \texttt{PASS}). \\
    \midrule
    \textbf{Capacitor Voltage Derating} 
    & Buck \texttt{CIN} ($36\,\text{V}$) \newline Buck \texttt{COUT} ($12\,\text{V}$) 
    & \texttt{T491A476K010AT}: Rated \textbf{$10\,\text{V} < 36\,\text{V}$ bus}. \newline \texttt{GRM32EC71A107KE15L}: Rated \textbf{$10\,\text{V} < 12\,\text{V}$ out} (Severe overvoltage conflict). 
    & \texttt{C44AHFP5300ZC0J}: Rated \textbf{$600\,\text{V} \gg 36\,\text{V}$}. \newline \texttt{EKXJ201ELL151MJ50S}: Rated \textbf{$200\,\text{V} \gg 12\,\text{V}$} (Sufficient safety margins). \\
    \midrule
    \textbf{Magnetic Realization} 
    & DAB \texttt{T1} \& \texttt{L\_SER} 
    & \texttt{DAB\_XFMR\_2\_1} \& \texttt{DAB\_LEAKAGE\_IND}: Unverified custom text placeholders. 
    & \texttt{HPH6-0325L} (4P:2S winding config.) \& \texttt{ATCA-01-101M-V}: Verified COTS magnetics. \\
    \midrule
    \textbf{Controller Role \& Ratings} 
    & Buck \texttt{CONTROL} 
    & \texttt{TPS54331DDCR}: $28\,\text{V}_{\max}$, $570\,\text{kHz}$ integrated-FET IC misused for $36\,\text{V}$, $80\,\text{kHz}$ external FET. 
    & \texttt{UCC28C53DR} + \texttt{UCC27714D}: Verified $80\,\text{kHz}$ current-mode PWM \& gate driver. \\
    \midrule
    \textbf{Power-Stage Connectivity} 
    & Buck \& DAB Netlists 
    & Input/output nodes shunted only to caps; shorted gate sources (\textbf{Singular matrix, 0\% sim pass}). 
    & Complete power-stage \& driver loop (\textbf{100\% convergence \& spec pass}). \\
    \bottomrule
  \end{tabular}
  \label{tab:failure_cases}
\end{table}

% ==========================================
% Section 5: Discussion and Limitations
% ==========================================
\section{Discussion and Limitations}
\label{sec:discussion}

The experimental results and comparative audits demonstrate that combining domain-specific power electronics synthesis, evidence-backed component verification, and closed-loop simulation advances automated design from ``generating plausible part numbers'' to delivering traceable, executable engineering artifacts. Nevertheless, to maintain strict academic and engineering transparency, several scope boundaries and limitations of the current implementation warrant discussion:

\begin{enumerate}
    \item \textbf{Conservative Derating vs. Multi-Objective Pareto Optimization:} Our current basic selection engine prioritizes provenance verifiability, rating compliance, and connectivity completeness rather than global cost, volume, or efficiency optimization. Consequently, when querying finite local/distributor candidate pools under strict contract constraints, the engine may select conservatively over-rated components—for instance, assigning $600\,\text{V}$ MOSFETs (\texttt{IPW60R041P6FKSA1}) and $200\text{--}600\,\text{V}$ film/electrolytic capacitors in the $24\text{--}48\,\text{V}$ Buck and DAB cases (yielding safety ratios of $7.8\times$ to $16.1\times$ above minimum derated thresholds). While electrically safe for functional prototyping, future extensions will incorporate multi-objective cost, footprint, and switching-loss ($Q_g \times R_{DS(\text{on})}$) Pareto ranking.
    \item \textbf{Simulation Model Fidelity Hierarchy and Closed-Loop Control:} In our default verification path, time-domain \texttt{ngspice} waveforms validate externally modulated power stages and source-derived gate-driver adapters using parameterized behavioral semiconductor models, while modular controller hardware (e.g., \texttt{UCC28C53DR}, \texttt{F280039CSPM}) is synthesized separately for schematic connectivity. Thus, passing power-stage reference simulation does not constitute full-system closed-loop stability proof, DSP firmware validation, or electro-thermal Safe Operating Area (SOA) qualification.
    \item \textbf{Sample Size and Component Coverage Gaps:} Even within the full framework, certain specialized passive positions (such as the Buck inductor under strict current-saturation evidence requirements in the comparative audit) may remain unverified (\texttt{UNKNOWN}) when local datasheet evidence is incomplete, highlighting the need for broader distributor API and multimodal datasheet coverage. Furthermore, while the two-case comparison clearly exposes structural and parametric failure modes of one-shot LLMs, expanding the benchmark to larger query sets and multi-turn tool-assisted baselines remains an important direction for future statistical evaluation.
\end{enumerate}

% ==========================================
% Section 6: Conclusion and Future Work
% ==========================================
\section{Conclusion and Future Work}
\label{sec:conclusion}

In this paper, we presented a \textbf{Bounded Dual-Loop Automated Schematic Synthesis Framework} that translates unstructured natural language specifications into verified, editable power electronics hardware deliverables. Rather than relying on open-loop text generation, our workflow coordinates domain-knowledge-driven circuit synthesis (covering Buck, Boost, Flyback, DAB, and half-bridge LLC topologies), contract- and evidence-backed commercial component selection, and closed-loop \texttt{ngspice} verification with stress feedback, culminating in modular control hardware integration and native \texttt{KiCad} netlist round-trip verification. End-to-end demonstrations on a $24\text{--}36\,\text{V}$ to $12\,\text{V}/2\,\text{A}$ Buck converter (29 positions, 83 verified terminals) and an isolated $48\,\text{V}$ to $24\,\text{V}/0.5\,\text{A}$ DAB converter (87 positions, 341 verified terminals) confirm the framework's ability to produce synchronized design reports, production BOMs, SPICE netlists, and 100\% connectivity-verified schematics. Comparative evaluations against a one-shot Vanilla LLM baseline further demonstrate improvements in component evidence verification from $22.2\%$ to $98.6\%$, datasheet nominal-value consistency from $0.0\%$ to $97.7\%$, and SPICE simulation pass rates from $0.0\%$ to $100.0\%$.

This work demonstrates that reliable AI-assisted hardware engineering lies in orchestrating semantic model comprehension with symbolic domain solvers, real-world supply-chain evidence, and physical simulators. Building upon the verification boundaries established in this study, our future work will focus on two primary directions:
\begin{enumerate}
    \item \textbf{Multi-Objective Component Optimization and High-Fidelity Electro-Thermal Verification:} To move beyond conservative derating margins and behavioral switch approximations, we plan to integrate real-time pricing, footprint constraints, and semiconductor loss figures-of-merit ($R_{DS(\text{on})} \times Q_g$) for Pareto-optimal component selection, while binding vendor-specific nonlinear SPICE subcircuits to qualify Safe Operating Area (SOA) and soft-switching (ZVS/ZCS) boundaries.
    \item \textbf{Closed-Loop Control Synthesis and Layout-Aware PCB Realization:} We aim to extend the modular control stage from pin-level hardware connectivity to automated small-signal compensator sizing and digital control firmware generation, while coupling custom magnetic design with automated KiCad PCB layout synthesis to close the loop between schematic verification and parasitic-aware physical implementation.
\end{enumerate}

%Bibliography
\bibliographystyle{unsrt}  
\bibliography{references}  

\end{document}